# MFORT-QA: Multi-hop Few-shot Open Rich Table Question Answering

Che Guan, Ph.D.
AllianceBernstein
Nashville, TN, USA
che.guan@alliancebernstein.com

Mengyu Huang
Snowflake
USA
renee.huang.823@gmail.com

Peng Zhang, Ph.D.
Vanderbilt University
Nashville, TN, USA
peng.zhang@vanderbilt.edu

## ABSTRACT

In today's fast-paced industry, professionals face the challenge of summarizing a large number of documents and extracting vital information from them on a daily basis. These metrics are frequently hidden away in tables and/or their nested hyperlinks. To address this challenge, the approach of Table Question Answering (QA) has been developed to extract the relevant information. However, traditional Table QA training tasks that provide a table and an answer(s) from a gold cell coordinate(s) for a question may not always ensure extracting the accurate answer(s). Recent advancements in Large Language Models (LLMs) have opened up new possibilities for extracting information from tabular data using prompts. In this paper, we introduce the Multi-hop Few-shot Open Rich Table QA (MFORT-QA) approach, which consists of two major steps. The first step involves Few-Shot Learning (FSL), where relevant tables and associated contexts of hyperlinks are retrieved based on a given question. The retrieved content is then used to construct few-shot prompts as inputs to an LLM, such as ChatGPT. To tackle the challenge of answering complex questions, the second step leverages Chain-of-thought (CoT) prompting to decompose the complex question into a sequential chain of questions and reasoning thoughts in a multi-hop manner. Retrieval-Augmented Generation (RAG) enhances this process by retrieving relevant tables and contexts of hyperlinks that are relevant to the resulting reasoning thoughts and questions. These additional contexts are then used to supplement the prompt used in the first step, resulting in more accurate answers from an LLM. Empirical results from OTT-QA demonstrate that our abstractive QA approach significantly improves the accuracy of extractive Table QA methods.

## KEYWORDS

Information Retrieval, Open Table Question-Answering, Large Language Models, Few-Shot Learning, Chain-of-Thought Prompting

## 1 INTRODUCTION

In today's rapidly evolving world, the power of artificial intelligence technology has brought significant advancements to a range of traditional industries such as healthcare, legal, and financial services. However, even with these remarkable strides, industry experts continue to face a pressing challenge in their daily work: the need to sift through vast amounts of information to extract important metrics. Often these vital metrics are embedded in table cells as hyperlinks, posing a significant obstacle to efficient data extraction and utilization.

To address this challenge, Table Question Answering (QA) has emerged as a solution that intelligently selects relevant tables from a collection and extracts answers to user queries directly from these tables. While modern Table QA training tasks employ tables with answers located in specific cell coordinates, this approach does not always guarantee accurate responses to complex queries. This difficulty is amplified when the answer is nested within a context of a hyperlink associated with a table, particularly if the table contains a large number of embedded hyperlinks. For example, within a table of a Wikipedia article, there can be many hyperlinks that include additional passages describing content in the table cells. In this case, it can be quite difficult to retrieve candidate tables along with the most relevant hyperlinked context due to the large size of the combined context. Additionally, language models may struggle to comprehend the intricacies of complex questions, further complicating the entire process. For instance, to answer this complex question, "Who created the series in which the character of Robert, played by actor Nonso Anozie, appeared?" [2], a model would need to identify the particular series in which the character of Robert, portrayed by actor Nonso Anozie, was featured. Answering this question requires the model to recognize the specific show and the creative individual responsible for its development.

In this paper, we propose a novel approach of Multi-hop Few-shot Open Rich Table Question Answering (MFORT-QA) that leverages an Open Table and Text QA dataset (OTTQA) [2] to more accurately answer complex questions. The main contributions of this paper include:

(1) Pioneering the use of few-shot learning (FSL) with LLMs for precise and succinct answers in open table QA.
(2) Employing chain-of-thought (CoT) to decompose complex questions into sub-questions with reasoning thoughts, and using retrieval augmented generation (RAG) to retrieve additional context for these sub-questions with reasoning thoughts. This enhances the initial prompt and assists LLMs in generating the final answer step-by-step.

The key idea of FSL [1] in this research work is to retrieve relevant tables for a given question from the test dataset and use them to generate accurate answers. To address the challenge of nested answers within hyperlinks in tables, we retrieve and append top k most relevant hyperlinked passages to each candidate table. We then use the question and candidate table to find similar question-table samples with actual answers from the training dataset. The retrieved ternary pair(s) of question-table-answer serve as a few-shot prompt in an LLM model, enabling us to leverage the full understanding capabilities of the LLM model and generate accurate answers from the retrieved tables.

To tackle the challenge of complicated questions, we employ the same few-shot prompt strategy to extract answer(s) from the candidate table(s) directly. If the answer cannot be generated, we apply CoT [23] to decompose the original questions into a few simpler sub-questions by using an LLM to generate a chain-of-thought reasoning and sub-questions in a sequential manner. Next, we utilize the RAG algorithm [6] by employing information retrieval techniques to gather additional relevant table information for these sub-queries to complement the prompt in the first step. This multi-hop process provides retrieved augmented information to assist the LLM in generating a precise answer. The experimental results demonstrate that the proposed approach achieves significantly improved accuracy compared to traditional extractive table QA methods.

In our exploration of the Table QA domain, we have reviewed several state-of-the-art fine-tuned QA models and their associated datasets, namely TableBert [3], SAT [27], SaMoE [30], and PASTA [7]. A common thread among these models is their extractive approach to QA tasks. Specifically, they leverage the input question in tandem with a table provided during the querying phase to predict the answer span directly from the supplied table content. This approach contrasts with our methodology, which introduces a novel strategy combining FSL, CoT and RAG. Notably, our baseline model utilizes the RoBERTa [15], a variant of the BERT architecture, to pinpoint the answer span within the table. Given the prevailing extractive trend in existing models, we posit that our RoBERTo-based baseline serves as an apt and representative benchmark, enabling a robust evaluation and comparative analysis against other Open Table QA models.

## 2 RELATED WORK

The recent surge in Large language models (LLMs) through pre-training Transformer models on extensive corpora has shown impressive capabilities in solving various natural language processing tasks. In particular, when the size of a model's parameters surpasses a certain threshold, these LLMs not only achieve a considerable improvement in performance but also exhibit certain special abilities (e.g., in-context learning), which are absent in small-scale language models (e.g., BERT) [28]. Several notable examples highlight the diverse capabilities of LLMs. Chinchilla [10] and PALM [5] employ parameter-efficient and adaptive approaches to enhance performance on tasks with limited resources. Gopher [18] focuses on cross-lingual transfer learning, while ChatGPT is specifically designed as conversational AI [17, 19]. The GPT-4[16], a large-scale and multimodal model which can accept image and text inputs and produce text outputs and has even more advanced capabilities than its predecessor. On the other hand, LLaMA2 [21], known as Open and Efficient Foundation Language Models, encompasses a comprehensive set of foundation language models with parameter sizes ranging from 7B to 65B. By careful attention design, Mistral 7B [12] achieve high performance while maintaining efficient inference, The model claims to outperform the LLaMA 2 -13B model across all tested benchmarks. These LLMs represent state-of-the-art machine learning models with exceptional understanding capabilities and the ability to excel in a wide range of NLP tasks, including providing accurate answers to user questions.

One popular downstream task of applying LLMs is question answering, especially Table QA, which attracts much attention in industries such as healthcare, legal, and financial services. Early Table QA models adapted the sequence-to-sequence (seq2seq) models to map natural language questions to SQL queries, such as Neural Enquirer [25] and Seq2SQL [29], allowing the retrieval of answers from structured data. The introduction of the Transformer architecture [22] revolutionized NLP and led to a series of breakthroughs in Table QA. Transformers are characterized by self-attention mechanisms that allow the model to weigh the importance of different input elements when making predictions. Notable models in this category include SQLova [11], TaBERT [26], TAPAS [9], and TAPEX [14]. More recently, research in Table QA has started to explore multi-modal approaches, such as HybridQA [4], which combine textual and structured data representations, aiming to better capture the inherent nature of table data. While these extractive Table QA models strive to provide precise answers from tables to address a user's query, they may be limited in their ability to handle table and nested hyperlink context in order to provide accurate answers to complex questions. RAG can thus be helpful to retrieve augmented context to aid an LLM in enhancing performance.

## 3 APPROACH

To demonstrate the approach of MFORT-QA, we can refer to an example illustrated in Figure 1. In this example, simply applying ChatGPT alone is insufficient to answer the initial question: "Who created the series in which the character of Robert, played by actor Nonso Anozie, appeared?". However, by decomposing the complex question into two sub-questions, these simple sub-questions along with reasoning thoughts aid the approach in obtaining retrieval augmented generations and further facilitate ChatGPT in producing the correct answer "Lynda La Plante" in the end.

To illustrate the two-step process employed by MFORT-QA for generating an answer(s) to a question, we can refer to Figure 2. In the stage of FSL, MFORT-QA evaluates the input question's embedding by comparing it to embeddings of annotated tables from a development set. This helps identify relevant tables that likely contain the answer. The model computes the embedding of the question and candidate table pair and contrasts it with embeddings of previously annotated examples from a training set. Relevant question-table-answer triples are chosen as few-shot prompts, which are then fed to ChatGPT to generate an initial response.

However, complex questions and tabular data with hyperlink contexts can impede the generation of an accurate answer. To address this, the second stage incorporates CoT prompting and RAG. The CoT prompting breaks down the initial question into simpler sub-queries along with reasoning thoughts. Then, RAG is used to retrieve relevant tables and hyperlink passages for both sub-queries 1 and 2. Finally, the initial question, the candidate tables, relevant support examples obtained from FSL, as well as supplement contexts retrieved from CoT and RAG are fed to ChatGPT to produce an answer. Both FSL, CoT and RAG are important for MFORT-QA to generate high-quality answers for complex questions involving tables.

| Question | | Who created the series in which the character of Robert, played by actor Nonso Anozie, appeared? |
|---|---|---|
| Few-Shot Learning | ChatGPT | Not answerable |
| Hop 1 | Sub-Question 1 | Which series it is that the character of Robert, played by actor Nonso Anozie, appeared? |
| | Answer to Sub-Question 1 | Prime Suspect 7: the Final Act (series name that Nonso Anozie played Robert) |
| Hop 2 | Sub-Question 2 | Who created the series Prime Suspect 7: The Final Act? |
| | Answer to Sub-Question 2 | Prime Suspect 7 is directed by Lynda La Plante |
| Final Answer | | Lynda La Plante |

**Figure 1: Illustration of the Proposed Model MFORT-QA Workflow: Example Question Answering Process**

Detailed explanations of each step will be provided in the subsequent subsections. In subsection 3.1, we begin by introducing general Table based QA tasks along with our choice of implementation. We then delve into the specifics of FSL for enhancing answer generation in subsection 3.2. Additionally, in subsection 3.3, we elaborate on the CoT approach, which effectively addresses complicated questions, and RAG, which effectively supplements relevant contexts for the initial prompt.

## 3.1 Table based QA Tasks

General QA is composed of two main components: the Information Retrieval (IR) system and the Reading Comprehension (RC) system. The IR system focuses on extracting relevant documents, passages, or tables from a large corpus of text that might contain the answer to a question. This retrieval is typically achieved using search engines or database query languages, which search for documents or passages that contain the question's keywords or phrases. Once a potential set of answers has been identified by the IR system, the RC system takes over. The RC system analyzes the retrieved content (documents, passages, and tables) to identify and extract the most relevant answer. This is typically done using pre-trained language models, such as BERT, which are able to understand the meaning of text and identify the most important information. These two systems work collaboratively to ensure the accuracy and efficiency of a QA system. The effectiveness of both the IR and RC systems is crucial in delivering precise answers to questions. In this study, our main focus is on Table-based QA tasks. As depicted in Table 1, these tasks can be classified into four distinct categories:

(1) **Table QA** answers questions over structured tables of data, where the answer can be found directly in a table cell.

**Table 1: Table QA Classification**

| Task | Table Given | Task Specific Reader Training | Task Specific Retriever Training |
|---|---|---|---|
| Table QA | Yes | Yes | N/A |
| Open Table QA | No | Yes | Maybe |
| Few-Shot Table QA | Yes | No | N/A |
| Few-Shot Open Table QA | No | No | Maybe |

(2) **Open-domain Table QA** uses a retriever to search for relevant tables in a large collection of table data and then trains a reader to extract answers from those tables.
(3) **Few-shot Table QA** answers questions over structured tables of data with a few training examples in prompt.
(4) **Few-shot Open-domain Table QA** prompting a general-purpose auto regressive language model with a few training examples to generate an answer in response to a question, without explicit training for answering questions or confirming if the response corresponds to an answer.

The traditional Open Table QA methodology typically involves using a pre-trained model like BERT to predict the gold cell as the answer from a table. However, this approach may not be suitable for rich tables that contain passages from hyperlinks. One solution is to flatten the table as text and append hyperlinked passages to the flattened table as rich table context. Traditional Text QA models can then be applied to predict the answer span from this enriched context. Alternatively, we can feed this rich table context directly to ChatGPT and leverage its understanding capability with FSL to prompt ChatGPT to generate an answer.

To implement the IR component, we utilize Sentence-BERT [20] to generate embeddings for flattened tables obtained from the OTT-QA dataset. These embeddings are then indexed in a vector database. During the retrieval process, we select the top candidate tables based on the cosine similarity between the candidate embedding and the question embedding. The tables with higher cosine similarities are considered as top candidates.

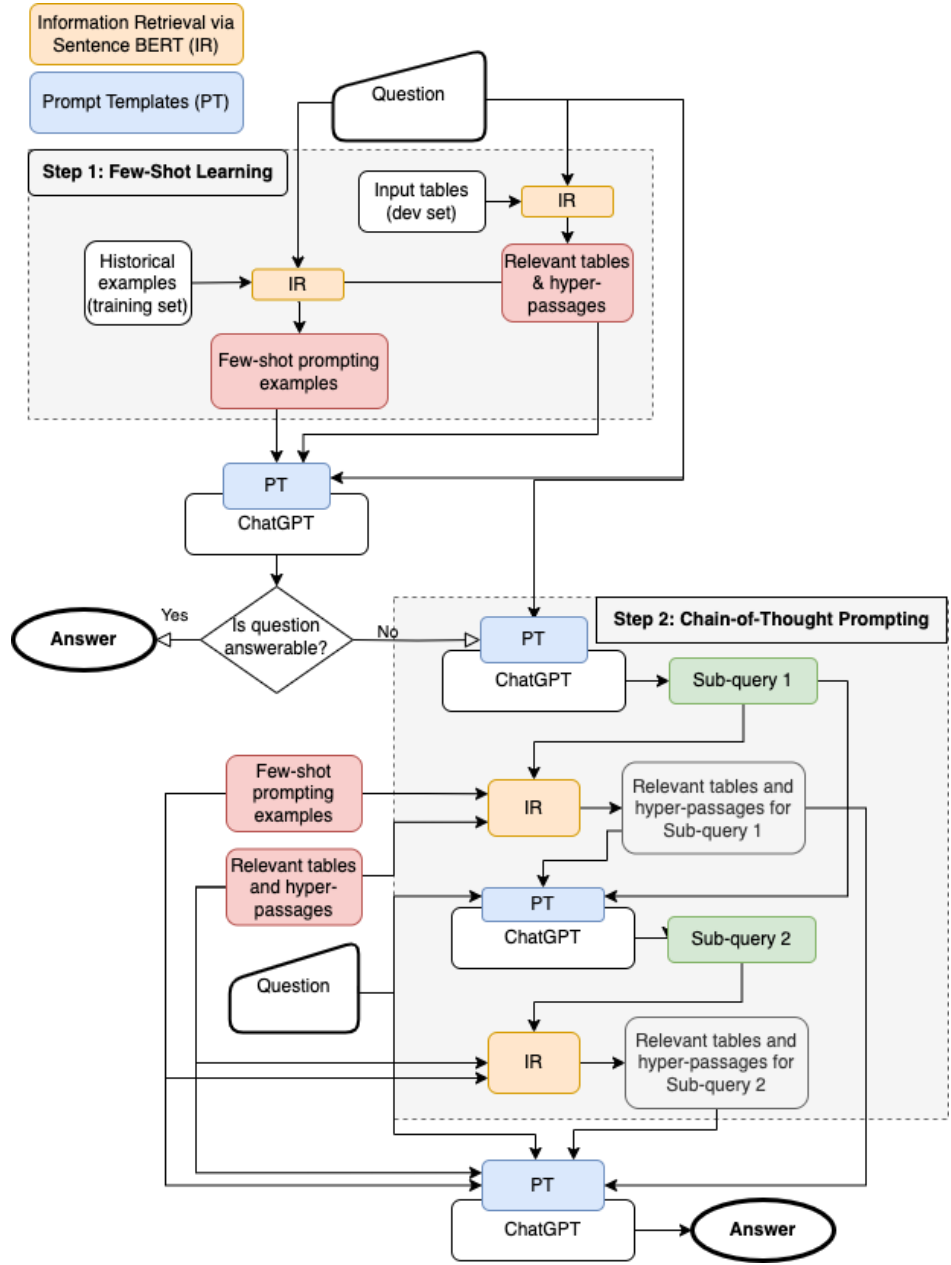

Figure 2: MFORT-QA Flowchart

To implement the RC component, we evaluate the performance of pre-trained language models such as TAPAS and RoBERTa. Additionally, we incorporate ChatGPT to potentially enhance the system's performance. Our IR methodology consists of two main stages. In the initial stage, we identify candidate tables from the testing set that may contain answers to the given question. Subsequently, we retrieve a few relevant samples from the extensive training set. These samples serve as valuable few-shot examples within our prompt template. If we cannot generate a correct answer to the initial question, in the second stage, we utilize CoT to break down complex questions into simple ones, and leverage RAG to complement the relevant contexts pertaining to the question. It is important to note that the relevance of the retrieved content to the question significantly improves the accuracy of the answer.

### 3.2 FSL

The first step of FSL involves using cosine similarity to retrieve a set of candidate tables from the testing set that may contain the correct answer(s) to the question. To achieve this, an encoder model like Sentence-BERT is used to embed the question and all the tables

Table 2: The Casts of the Principal Original Productions

| Role | Opera Comique, 1877 | Savoy Theatre, 1884 |
|---|---|---|
| Sir Marmaduke | Richard Temple | Richard Temple |
| Alexis | George Bentham | Durward Lely |
| ... | | |
| Mrs. Partlet | Harriett Everard | da Dorée |
| Oldest Inhabitant | Frank Thornton | - |

Table 3: List of Jewish Actors

| Name | Years | Nationality/profession | References |
|---|---|---|---|
| Gertrude Berg | 1899-1966 | American radio and television actress | [743] |
| ... | | | |
| Hermione Gingold | 1897-1987 | Austrian film actress | [748] |

separately. Based on the similarity scores, the top-k candidate tables are ranked and selected.

To construct Few-shot examples for prompting ChatGPT, the initial question is combined with each candidate table to calculate their respective embeddings. Furthermore, each ternary pair of Question-Table-Answer from the training set is combined and its embedding is computed. Using cosine similarity once more, the top ternary pairs of Question-Table-Answer are chosen as few-shot examples to be included in the prompt.

Finally, the initial question, the top candidate tables derived from the testing set, and the selected few-shot support pairs of Question-Table-Answer from the training set are provided as input to ChatGPT. These examples are utilized to facilitate the generation of an answer to the initial question by ChatGPT. To illustrate the process of FSL with one-shot learning, the example below outlines the steps involved:

**Initial Question**: Which Mrs. Partlet actress from the comic opera The Sorcerer died at the age of 37?

**Step 1**: Retrieve a candidate table as shown in Table 2, which is relevant to the initial question from testing set, using the cosine similarity:

**Step 2**: Retrieve a question-table-answer pair from the training set that is relevant to both (1) the initial question and (2) the candidate table above, using cosine similarity. This will provide the employed LLM with an example of the question-answering format and style, which can be used as a one-shot prompting template.

- **Example Question**: In which year did an accident end the career of the Jewish British actress who lived from 1897-1987?
- **Most Relevant Table** for Example Question: Table 3
- **Most Relevant Hyperlinked Passage** from Reference [748]: 'From the early 1950s, Gingold lived and made her career mostly in the U.S. ... She made appearances in revues and toured in plays and musicals until an accident ended her performing career in 1977'
- **Answer** for Example Question: 1977

**Step 3**: Generate the answer to the initial question using FSL: **Harriett Everard**

### 3.3 Multi-hop CoT and RAG

The OTT-QA dataset is made up of unprocessed Wikipedia tables and corresponding passages obtained from embedded hyperlinks. In cases where answers cannot be found explicitly within the dataset, a logical approach is to derive a multi-step solution, although this process can pose challenges.

Drawing inspiration from [13], when faced with a complex question from a vast and complex relevant context, our study employs the CoT prompting that entails breaking down the original question into simpler sub-queries, referred to as "hops." The first sub-query can be directly generated from slightly modifying the initial FSL template. Starting with the generated sub-query 1, the IR procedure is then applied to find relevant tables and hyper passages associated with this sub-query 1. These tables, along with the original question and sub-query 1, are used as input to ChatGPT to generate sub-query 2. Relevant tables and hyper passages for sub-query 2 are subsequently retrieved.

Finally, using the RAG procedure, the retrieved tables and passages for both sub-queries 1 and 2 are combined and added to supplement the initial model prompt input. ChatGPT is then run again, taking into account the additional information obtained from this RAG procedure.

## 4 EXPERIMENTS

The performance of MFORT-QA was assessed through experimentation on the OTT-QA dataset, which was originally presented in the paper by Chen et al.,2021 [24]. The OTT-QA dataset comprises open-ended questions that require retrieving tables and text from the internet to generate answers. It is a re-annotation of the previous HybridQA dataset [4].

We present five cases to illustrate the effectiveness of our method. Case 1 demonstrates the value of information retrievals via Sentence-BERT model [20]. Case 2 showcases the performance improvement achieved by applying FSL prompting to the abstractive LLM, compared to an extractive BERT model. Case 3 illustrates the additional value obtained through CoT Prompting and RAG. Case 4 presents a specific example of how RAG retrieves augmented context and further enhances model performance. Case 5 exemplifies a step-by-step enhancement demonstrated in the Ablation Studies.

In all cases except Case 4, we utilized the data split from OTT-QA [24], where the training set (14,541 tables) and validation set (3,012 tables) were employed for conducting experiments. The performance evaluation encompassed various metrics, including hit rates for top-k retrievals as denoted as HITs@K, F1 Score, Recall, Precision, and Exact Match (EM).

### 4.1 Investigate IR

In this experiment, we use Sentence-BERT to embed flattened tables and questions, and use cosine similarity metric to evaluate the relevance. To ensure a proper number of tables and associated hyperlink passages are retrieved, besides the original dataset, we also create a filtered dataset by filtering the test set to only include tables with a direct answer to a given question other than answers that exist in associated hyperlink passages. We then apply Sentence-BERT to retrieve the top 1 and 3 most relevant tables separately. We evaluate the retrieval accuracy using the Hit rate metric (i.e., how many correct tables are retrieved as relevant tables) for top K retrievals and report the results for both the original whole test set and the filtered dataset.

Table 4: Retrieval Accuracy (%)

| Metrics | Filtered Data | Original Data |
|---|---|---|
| HITS@1 | 88.24 | 87.53 |
| HITS@3 | 96.24 | 96.16 |

Table 5: The Performance (%) of MFORT-QA with FSL on the Filtered Dataset

| Models | F1 | Precision | Recall | EM |
|---|---|---|---|---|
| Baseline | 21.32 | 19.99 | 26.32 | 13.48 |
| FSL | 29.59 | 28.35 | 34.20 | 24.32 |

Table 6: The Performance (%) of MFORT-QA with FSL on the Original Dataset

| Models | F1 | Precision | Recall | EM |
|---|---|---|---|---|
| Baseline | 8.12 | 8.46 | 8.81 | 6.32 |
| FSL | 25.61 | 25.36 | 29.97 | 19.24 |

Table 7: The performance (%) of MFORT-QA with both FSL and CoT&RAG on the filtered dataset

| Models | F1 | Precision | Recall | EM |
|---|---|---|---|---|
| Baseline | 21.32 | 19.99 | 26.32 | 13.48 |
| FSL+CoT&RAG | 33.90 | 32.57 | 39.68 | 27.95 |

Table 8: The Performance (%) of MFORT-QA with Both FSL and CoT&RAG on the Original Dataset

| Models | F1 | Precision | Recall | EM |
|---|---|---|---|---|
| Baseline | 8.12 | 8.46 | 8.81 | 6.32 |
| FSL+CoT&RAG | 32.35 | 31.88 | 38.21 | 24.24 |

Numerical results from Table 4 indicate that a more than 9% increase in HITs@K for both the filtered dataset and the original complete dataset when k is increased from 1 to 3, respectively. This comparison provides valuable insight into the importance of maintaining a larger number of pertinent tables and hyperlink passages. Consequently, the default parameters for the subsequent experiments are chosen as the top three tables and their corresponding hyperlink passages separately.

### 4.2 Investigate FSL

In this experiment, we first compare traditional Table QA to ChatGPT on the filtered dataset. We use Sentence-BERT to retrieve the top relevant table and employ TAPAS [9], a Table QA model, to predict the gold cells as a baseline. We then compare the performance of TAPAS and FSL of our approach only on the filtered dataset.

However, as TAPAS is not designed to handle tables with hyperlink passages, to ensure a fair comparison, we use Sentence-BERT to retrieve the most relevant table from the original dataset. We convert this table into flattened text and combine it with the top relevant hyperlinked passages for each flattened table. We then utilize Text QA: RoBERTa [15] to predict the answer span, treating it as a baseline. On the other hand, for our approach, we directly provide this enriched context to the FSL and construct a few-shot prompt for ChatGPT to generate an answer. This methodology allows us to compare the performance of our approach against the baseline.

The results as presented in Table 5 show that the model performance via FSL prompting on the filtered dataset improves 8% on F1 and 11% on EM in comparison to the baseline, respectively. Furthermore, the results as reported in Table 6 show that FSL prompting on the original dataset improves at least three times in comparison to the baseline.

### 4.3 Investigate Multi-hop CoT and RAG

FSL can effectively prompt ChatGPT to generate better results than the baseline. However, in situations where the question or table context is complex and difficult to understand, FSL alone may not be sufficient. To address this, the CoT procedure can be used to break down the difficult question into sequential simpler questions through Multi-Hops. RAG then utilizes these sub-questions to retrieve additional relevant tables and hyperlink passages to supplement the original input context and prompt ChatGPT.

In this experiment, similar to the results presented in Tables 5 and 6, we add the CoT and RAG on the top of FSL. The results as presented in Table 7 show that the model performance via CoT and RAG on the filtered dataset improves 12% on F1 and 14% on EM in comparison to the baseline, indicating a marginal improvement in comparison to applying FSL only. Furthermore, the results as reported in Table 8 show that CoT and RAG on the original dataset improves at least four times in comparison to the baseline, indicating around 30% improvement in comparison to applying FSL only.

### 4.4 Further Investigate RAG on an Illustrative Example

In the given illustrating example as shown in Figure 1, we have observed that MFORT-QA is capable of generating the correct answer to the question "Who created the series in which the character of Robert, played by actor Nonso Anozie, appeared?". Now, let's compare MFORT-QA with other popular QA methods on the same illustrative example.

**Table QA on Relevant Tables -> No Answer:** The objective of Table QA models is to extract answers directly from tables. In our scenario, we retrieve the most relevant tables using the given question. However, since the answer to our example question is embedded within a hyperlinked passage rather than being directly available in the most relevant tables, such as Table 9 obtained from the OTTQA dataset [24] using the Sentence-Bert model, even a Table QA model like TAPAS[8], which was selected for this experiment, cannot provide an accurate answer from any of the retrieved tables.

**Table QA on Relevant Tables + Text QA on Relevant Hyperlinked Passages -> No Answer.** As Table QA model TAPAS is

Table 9: Example of a Relevant Table

| Year | Title | Role | Notes |
|---|---|---|---|
| 2007 | Prime Suspect7: The Final Act | Robert | Episode: Part1 |
| 2009 | Occupation | Erik Lester | 3 episodes |
| 2011 | Outcasts | Elijah | 1 episode |
| 2011 | Stolen | Thomas Ekoku | TV movie |
| ... | | | |

unable to extract the correct answer, we then apply another modern text QA model, RoBERTa [15], to attempt answer extraction. We do so by first selecting the top 3 most relevant hyperlinked passages for each of the three retrieved tables. We then combine each hyperlinked passage with its table content to form a longer context that is inputted to RoBERTa. This combined Table and Text QA approach also resulted in no answer.

**Table QA on Relevant Tables + Text QA on Augmented Hyperlinked Passages retrieved from RAG -> Correct Answer:** In order to assess the effectiveness of RAG achieved through CoT in our proposed model, we divide the question into two sub-queries: "Series with character Robert played by Nonso Anozie?" and "Who created Prime Suspect 7: The Final Act?" Subsequently, we employ the same Text QA model RoBERTa as mentioned earlier on the hyperlinked passages retrieved after the second sub-query, leading to the successful extraction of the correct answer.

**ChatGPT (with Zero-Shot Learning) on combined Tables and Hyperlinked Passages retrieved from RAG -> Correct Answer:** By employing relevant combined tables and hyperlinked passages obtained from RAG as augmented context and utilizing a simplified prompt with zero-shot learning, we observe that ChatGPT successfully generates the correct answer when applying Text QA.

From the above comparison, we can not only observe the limitations of traditional extractive Table and Text QA models but also witness how RAG retrieve augmented context and further enhance model performance.

### 4.5 Ablation Studies of MFORT-QA

In order to conduct detailed ablation studies, we have incorporated two additional experiments. Firstly, we utilize the pure ChatGPT to answer questions on the original test dataset without utilizing IR. Secondly, we apply LLM with zero-shot learning while incorporating a retrieved candidate table using IR in the prompt to answer questions. These results are then compared to the outcomes obtained from the previous step, namely FSL and FSL+CoT&RAG

As shown in Table 10, this comparative analysis provides an illustrative demonstration of the step-by-step improvement achieved throughout the study. It can be observed that zero-shot learning significantly improves performance in comparison to applying pre-trained ChatGPT to answer questions. FSL with a few shot examples further improves performance, though marginally. The addition of CoT and RAG further adds an additional average 6% boost on top of FSL.

Table 10: Ablation Studies for MFORT-QA on the Original Dataset

| Ablation Studies | F1 | Precision | Recall | EM |
|---|---|---|---|---|
| LLM w/o Table | 3.36 | 3.24 | 4.39 | 2.12 |
| Zero-shot Learning | 23.09 | 24.05 | 23.46 | 17.63 |
| FSL | 25.61 | 25.36 | 29.97 | 19.24 |
| FSL+CoT&RAG | 32.35 | 31.88 | 38.21 | 24.24 |

## 5 CONCLUSION

We introduce MFORT-QA, a novel approach to multi-hop few-shot open rich table question answering. With respect to the IR procedure, retrieving the top three tables and their associated hyperlink passages produces a very high hit rate. ChatGPT, even with zero-shot learning, is effective for understanding tables and answering questions. Furthermore, ChatGPT with FSL enhances its understanding of retrieved examples for a given question in the prompt. This assists ChatGPT to generate accurate answers from the retrieved tables. For complex questions that ChatGPT struggles to answer directly, multi-hop CoT decomposes the question into simpler ones with reasoning thoughts. Leveraging this information, RAG further retrieves additional relevant content to supplement the initial prompt, significantly boosting the chances of generating accurate responses. Numerical testing results demonstrate that MFORT-QA with FSL, CoT, and RAG outperforms both traditional extractive table and text QA models and LLMs with zero-shot learning on the OTT-QA dataset.

As with any research endeavor, it is crucial to acknowledge the inherent limitations of our approach. In order to mitigate the noise in sub-questions, we refer to the 'DSP' approach [13]. However, the computational demands posed by OTT-QA dataset require careful consideration, presenting a challenge for directly adopting this approach. In future endeavors, it would be advantageous to explore open-source models like the LLaMA2 model [21] or Mistral-7B [12] , which can help minimize computation costs. Additionally, these models provide the probability of generated tokens. By leveraging the probability, for example through techniques like beam search, we may achieve further refinement of results. It is also important to note that the OTT-QA data used in our study originates from Wikipedia tables, which could potentially be part of the GPT3.5 pre-training corpus. Consequently, despite ChatGPT's prompts to source answers from the input text, there is a possibility that it may still draw upon its internal knowledge of Wikipedia. Acknowledging this potential influence is essential in interpreting the outcomes of our approach.